\documentclass[letterpaper]{article} 
\usepackage{aaai23}  
\usepackage{times}  
\usepackage{helvet}  
\usepackage{courier}  
\usepackage[hyphens]{url}  
\usepackage{graphicx} 
\usepackage{natbib}  
\usepackage{caption} 
\usepackage{algorithm}
\usepackage{algorithmic}
\usepackage{amsmath}
\usepackage{amssymb}
\usepackage{multirow}
\usepackage{subcaption}
\usepackage{makecell}
\usepackage{booktabs}

\usepackage{newfloat}
\usepackage{listings}
\DeclareCaptionStyle{ruled}{labelfont=normalfont,labelsep=colon,strut=off} 
\floatstyle{ruled}
\newfloat{listing}{tb}{lst}{}
\floatname{listing}{Listing}
\title{Learning to Learn Better for Video Object Segmentation}
\author {
    Meng Lan \textsuperscript{\rm 1},
    Jing Zhang  \textsuperscript{\rm 2},
    Lefei Zhang  \textsuperscript{\rm 1,4},
    Dacheng Tao \textsuperscript{\rm 3,2}
}
\affiliations {
    \textsuperscript{\rm 1} Institute of Artificial Intelligence and School of Computer Science, Wuhan University, China \\
    \textsuperscript{\rm 2} The University of Sydney, Australia \quad
    \textsuperscript{\rm 3} JD Explore Academy, China \quad
    \textsuperscript{\rm 4} Hubei Luojia Laboratory, China \\
     \{menglan, zhanglefei\}@whu.edu.cn, \  jing.zhang1@sydney.edu.au, \ 
    dacheng.tao@gmail.com
}

\usepackage{bibentry}

\begin{document}

\maketitle

\begin{abstract}
Recently, the joint learning framework (JOINT) integrates matching based transductive reasoning and online inductive learning to achieve accurate and robust semi-supervised video object segmentation (SVOS). However, using the mask embedding as the label to guide the generation of target features in the two branches may result in inadequate target representation and degrade the performance. Besides, how to reasonably fuse the target features in the two different branches rather than simply adding them together to avoid the adverse effect of one dominant branch has not been investigated. In this paper, we propose a novel framework that emphasizes Learning to Learn Better (LLB) target features for SVOS, termed LLB, where we design the discriminative label generation module (DLGM) and the adaptive fusion module to address these issues. Technically, the DLGM takes the background-filtered frame instead of the target mask as input and adopts a lightweight encoder to generate the target features, which serves as the label of the online few-shot learner and the value of the decoder in the transformer to guide the two branches to learn more discriminative target representation. The adaptive fusion module maintains a learnable gate for each branch, which reweighs the element-wise feature representation and allows an adaptive amount of target information in each branch flowing to the fused target feature, thus preventing one branch from being dominant and making the target feature more robust to distractor. Extensive experiments on public benchmarks show that our proposed LLB method achieves state-of-the-art performance.

\end{abstract}

\section{Introduction}
Semi-supervised Video Object Segmentation (SVOS) is a task of finding the pixel-level position of the target objects in a video sequence, given the mask annotation at the first appearance of the target. It has received great attention in recent years for its wide applications, such as video editing and video surveillance \cite{lu2021omnimatte}. While SVOS is an extremely challenging problem, since the target object annotation is only defined in the first frame and the object may undergo various appearance and scale changes in subsequent frames, which usually contain distractors in the background.

\begin{figure}[t]
  \centering
  \includegraphics[width=0.96\linewidth]{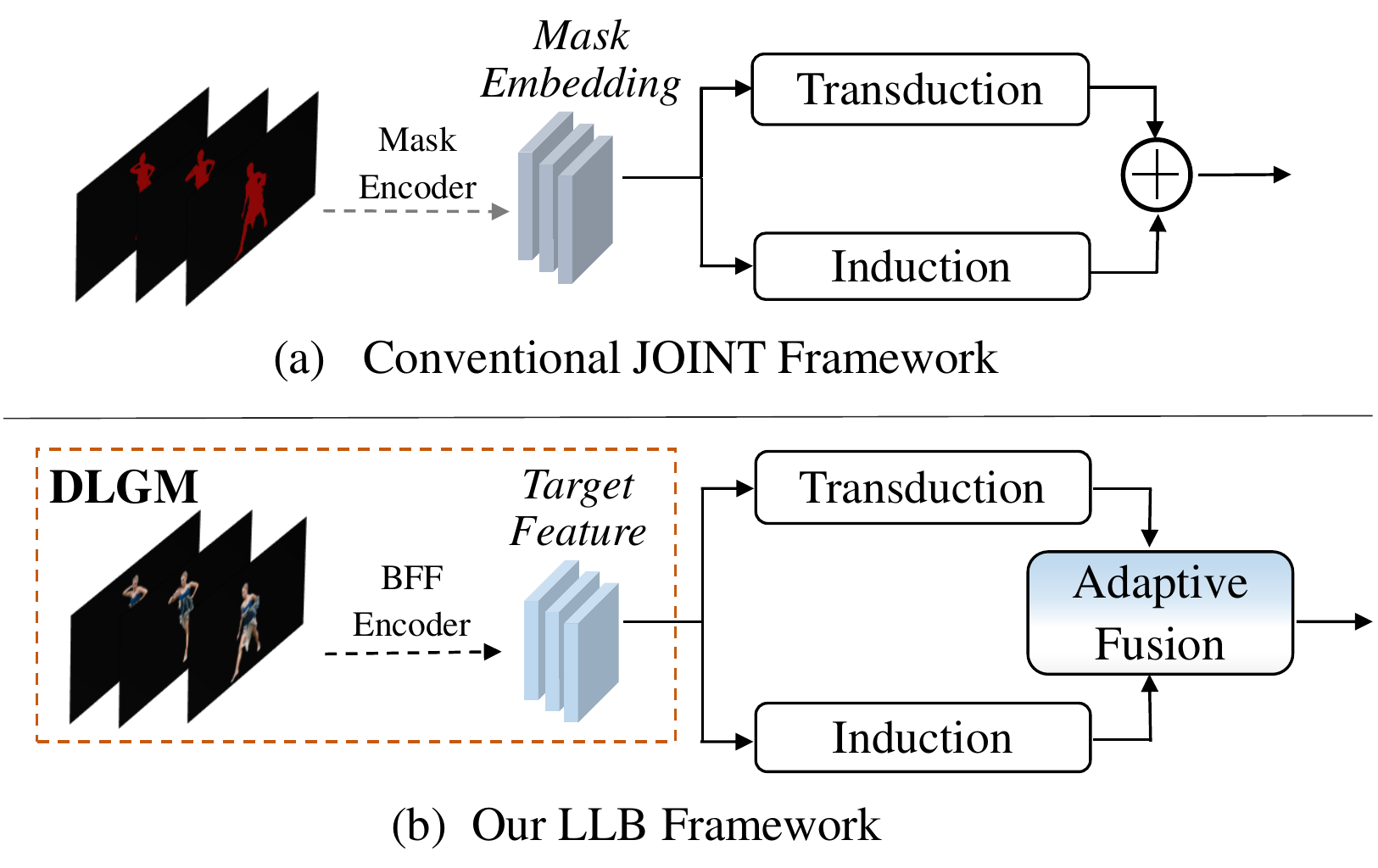}
  \caption{Comparisons of the conventional JOINT framework \cite{mao2021joint} and our proposed LLB framework. Compared with JOINT, our LLB model devises the DLGM to provide more discriminative target features for the two branches to learn better target representation. The adaptive fusion module reweighs the element-wise feature representation in each branch and then adaptively aggregates the target information from both branches, making the fused target representation more robust and discriminative.}
  \label{fig_1}
\end{figure}

In recent years, most of the advances in the SVOS field focus on how to capture and exploit target information, as feature extractor and segmentation decoder are almost similar. A mainstream direction is to employ the transductive reasoning strategy which performs the pixel-level matching between the reference frames and the current frame to realize the target information propagation from the past to the current segmentation. These matching based methods, e.g., STM \cite{STM} and HMMN \cite{HMMN}, have set state-of-the-art of SVOS and promoted the development of the field. On the other hand, the online few-shot learning based methods provide a new perspective of online inductive learning on SVOS and also achieve impressive performance \cite{FRTM,LWL}. Taking the representative LWL \cite{LWL} as example, a target model is learned to transform the general feature of current frame to an internal high-dimension target mask encoding, and a differentiable few-shot learner is trained online to predict the parameters of the target model, which uses the reference frame along with its predicted mask to adapt the model to the target object. 

To leverage the advantages of the offline transductive reasoning and the online inductive learning, JOINT \cite{mao2021joint} designs a two-branch architecture to jointly integrate a matching based transformer and an online few-shot learner within a unified framework, in which the transformer aggregates rich spatio-temporal information while the induction branch provides superior discrimination capability, as shown in Fig. \ref{fig_1} (a). This is an interesting and effective attempt and has achieved impressive performance. However, there are also some flaws in the current joint learning framework for SVOS. First, it inherits the mask encoding of the LWL \cite{LWL} to serve as the supervised label of the few-shot learner and the target cues of the past frames in the transformer, respectively. While the single-channel mask contains only the target position cues and has few target representation information, such as the color, texture and illumination intensity. This may lead the model to generate target feature with weak representation. Besides, as we can observe, most of the matching based approaches \cite{STM,LCM} encode the past frames and their labels as the memory rather than only the mask embedding. Second, the JOINT framework directly sums the output of the two branches, without considering the balance of the target information in the two branches and the potential distractor representation, which may lead to the dominance of one branch overwhelming the representation of the other branch.

To address these issues, we propose a novel framework that emphasizes learning to learn better target features for SVOS named LLB in this work, where we design a discriminative label generation module (DLGM) and an adaptive fusion module (AFM). As shown in Fig. \ref{fig_1} (b), in the DLGM, we multiply the mask with the corresponding frame to generate the background-filtered frame (BFF) and adopt a lightweight yet powerful encoder to encode the BFF to the target feature. Compared with the mask embedding, the BFF based feature enhances the foreground representation and the discriminative target feature can guide the online trained target model to predict better target feature in the induction branch and promote the propagation of the target information in the transduction branch. To better fuse the output target features of the two branches and suppress potential noise, the AFM learns a gate for each branch. The learnable gate adaptively reweighs the pixel-wise feature representation and controls the amount of target information in each branch that flows to the fused feature, making the final target feature more robust and discriminative. The experimental results on several benchmarks show that our LLB can significantly improve the performance of the existing joint learning based SVOS framework.

The contributions of this work can be summarized as follows:
\begin{itemize}
\item We propose a new joint learning based framework for SVOS task, which enhances the target representation of the past frames in both branches and adaptively fuses their output features for learning more robust target representation. The proposed model achieves state-of-the-art performance on the public challenging benchmarks.

\item We devise the discriminative label generation module to encode the BFF into discriminative target feature, which not only serves as the label of the few-shot learner to guide the target model to predict better target feature, but also works as the spatio-temporal context in the transformer to reinforce the propagation of target information from the past frames to the current segmentation.

\item To balance the learning of each branch and make the fused target feature more robust, we design the adaptive fusion module to learn a gate for each branch to reweigh the element-wise feature and allow adaptive amount of target information flowing to the fused target feature.

\end{itemize}

\section{Related Work}
In recent years, great progress has been made in the field of video object segmentation, leading to remarkable performance improvement. In the early age of SVOS, the methods \cite{OSVOS,PReMVOS} mainly adapted a general semantic segmentation network to the SVOS task through online fine-tuning, while this strategy easily leads to overfitting to the initial target appearance and impractical running time. Therefore, more recent methods integrate target-aware models into a target-agnostic segmentation framework, where the target models capture the target information from the reference frame-mask pairs and provide the generated target representation to the segmentation decoder to predict the final segmentation mask. Generally, most SVOS frameworks share similar feature extractors and segmentation heads, and the difference is how the target models utilize and encode the target information from the reference frames. Based on the learning mode of the target model, we classify these methods into two categories.

\subsection{Offline Learning Based Methods} This type of methods employs the transductive learning strategy and does not involve the online learning process during inference. It could be roughly grouped into two directions. 1) \textbf{Propagation based methods} implement the target model by introducing the target features from the historical frames to the current frame. For example, RGMP \cite{RGMP} concatenates features of the first, previous, and current frames to explicitly enhance the target representation for the segmentation head. SAT \cite{SAT} updates a dynamic global feature of the target and propagates it to the current frame. Although directly concatenating the target features of the reference frames is efficient for SVOS, the drifting remains a major problem between sequential frames. 2) \textbf{Matching based methods} are the mainstream research direction for their finer design and better performance. These methods find the target-specific information in the current frame by calculating the pixel-level similarity with the reference frames. Among them, STM \cite{STM} maintains a memory to store target information in reference frames and retrieves memory using non-local and dense memory matching to find the target object in the current frame. CFBI \cite{CFBI} combines the foreground and background matching together with an instance-level attention mechanism to learn a robust target feature. LCM \cite{LCM} proposes to integrate the pixel-level matching, the object-level information and the position consistency together to obtain a better target representation. AOT \cite{AOT} designs a long short-term transformer to construct hierarchical matching and propagation of the target information. RDE-VOS \cite{RDEVOS} propose the recurrent dynamic embedding to build a constant memory bank and an unbiased guidance loss to avoid error accumulation. These matching methods deliver the state-of-the-art performance. Since effective feature matching requires a powerful and generic feature embedding, these matching based approaches have to adopt the pre-training stage on large amount of synthetic video data to obtain better performance.

\subsection{Online Learning Based Methods} These methods always involve in the online learning process to adapt the model to the specific target during the inference, using the past frame-mask pair. For instance, EGMN \cite{EGMN} exploits an episodic memory network to store frames as nodes and capture cross-frame correlations by edges, and uses the learnable memory controllers to achieve online memory reading and writing. FRTM \cite{FRTM} designs a few-shot learner to predict the parameters of target model that achieve the transformation from the target-agnostic feature to a low-resolution target-specific mask. The target model in LWL \cite{LWL} is trained to predict a high-dimension mask encoding to contain rich target representation, which significantly improves the final segmentation performance.

Especially, JOINT \cite{mao2021joint} combines the advantages of the pixel matching transductive reasoning and the online few-shot learning into a unified framework to learn a high-quality target mask embedding. However, we argue that a mask cannot provide enough target cues to guide the two branches to learn more discriminative target feature and limit the performance of the whole framework. Moreover, JOINT directly sums the outputs of the two branches without considering the weight of each branch, which may result in the dominance of one branch and restrict the representation of another branch. In this paper, we propose the DLGM to produce discriminative target feature as the guidance of the two branch to generate better intermediate target representation and the adaptive fusion module to learn more robust and discriminative target feature for high-performance video object segmentation.

\begin{figure*}[t]
  \centering
  \includegraphics[width=0.92\linewidth]{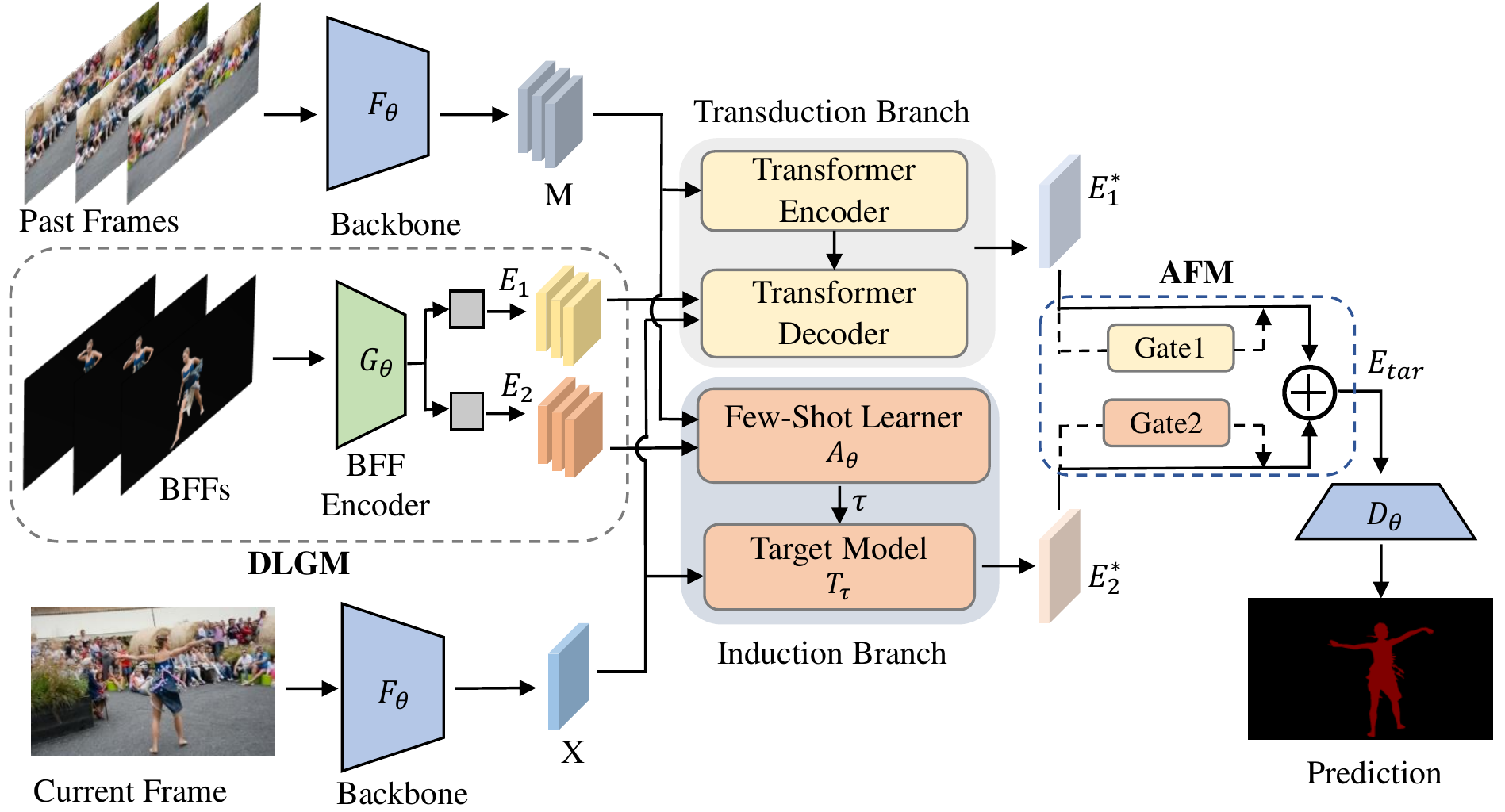}
  \caption{An overview of our LLB framework. The discriminative label generation module first multiplies the past frame-mask pairs to form the BFFs, and then encodes the BFFs into two complementary target encodings, which serve as the target spatio-temporal cues in the transduction branch and the supervised labels of the few-shot learner in the induction branch, respectively. The transduction branch retrievals fine-grained target information from the past frames via a lightweight transformer, while the induction branch trains an online few-shot learner to learn a target model to generate the target-aware feature. The adaptive fusion module learns a gate for each branch to reweigh the element-wise feature representation and then fuse the output features of the two branches to produce a more discriminative and robust target feature for the subsequent decoder.}
  \label{framework}
\end{figure*}

\section{Method}

\subsection{Overall Pipeline}
Following the architecture design of JOINT \cite{mao2021joint}, we build our new joint learning based framework for SVOS. As illustrated in Fig. \ref{framework}, LLB is composed of following parts: a deep feature extractor backbone $F_{\theta}$, a discriminative label generation module, a transformer based transduction branch, an online few-shot learner based induction branch, an adaptive fusion module, and a segmentation decoder $D_{\theta}$. During training, the past frames and the current frame are extracted by the backbone to obtain high-level image features $M \in \mathbb{R}^{N \times H \times W \times C}$ and $X \in \mathbb{R}^{H \times W \times C}$ respectively, where $N$ is the number of the past frames, $H$ and $W$ are the height and width, and $C$ is the channel dimension. The DLGM takes the past frame-mask pairs as input and output two target encodings $E_{1}$ and $E_{2}$ for the two branches. 

For the transduction branch, it adopts a lightweight transformer \cite{vit} to implement the propagation process and obtain the target aware feature. The transformer encoder takes the past frames features $M \in \mathbb{R}^{N \times H \times W \times C}$ as input, and the features are flattened into $\widetilde{M} \in \mathbb{R}^{NHW \times C}$ and then sent to the self-attention block, which consists of a self-attention layer, the residual connection, and instance normalization operations, to produce the feature $\widetilde{\mathbf{M}}_{\text{sa}} \in \mathbb{R}^{NHW \times C}$. The transformer decoder contains a self-attention block and a cross-attention layer. The self-attention block, the same as the one in the transformer encoder, takes the feature of the current frame $X$ as input and output the feature $\widetilde{X}_{sa} \in \mathbb{R}^{HW \times C}$. The cross-attention layer is the core operation to implement the propagation of the spatio-temporal information of target object from the past frames to the current frame. In the cross-attention layer, $\widetilde{X}_{sa}$ serves as the query, while the encoded past frame features $\widetilde{\mathbf{M}}_{\text{sa}}$ and the flattened target encoding $\widetilde{E}_{1} \in \mathbb{R}^{NHW \times D}$ are built as the key-value pair, which stores the target cues. The query retrieves the target information from the key-value pair and produces the target-aware feature $\mathbf{E}_{1}^{*} \in \mathbb{R}^{H \times W \times D}$.

In the induction branch, an internal few-shot learner $A_{\theta}$ is designed to predict the parameters $\tau$ of the target model $T_{\tau}$ by minimizing squared error on the training samples, which can be formulated as follows:
\begin{equation}
	L(\tau)=\frac{1}{2} \sum\left\|W_{\theta}\left(\hat{I}_{i}\right) \cdot\left(T_{\tau}\left(m_{i}\right)-G_{\theta}\left(\hat{I}_{i}\right)\right)\right\|^{2}\\+\frac{\lambda}{2}\|\tau\|^{2}.
\label{eq_3}
\end{equation}

Here, $m_{i}$ is the backbone feature of the past frame $I_{i}$ in the memory and $\hat{I}_{i}$ is the corresponding BFF. The target module $T_{\tau}$, as a differentiable linear layer, where $\tau \in \mathbb{R}^{ K \times K \times C \times D}$ is the weights of a convolutional layer with kernel size K, maps a C-dimension generic image feature to a D-dimension target-aware representation with the spatial size unchanged. Accordingly, the lightweight BFF encoder $G_{\theta}$ encodes the BFF to the same D dimension as the output of target model to provide the label supervision. $W_{\theta}(\hat{I})$ is the learnable position weights. After the training of the few-shot learner, the trained target model is applied to the current frame feature $X$ to generate the target representation $\mathbf{E}_{2}^{*} \in \mathbb{R}^{H \times W \times D}$.

The adaptive fusion module fuses the target features of the two branches and the fused target feature is further processed by a decoder to predict the final segmentation result.

\subsection{Discriminative Label Generation Module}
As shown in the Fig. \ref{fig_1} (a), the JOINT framework employs the mask embedding to guide the target-aware feature generation process in the two branches. However, the mask embedding encoded from a single-channel mask lacks rich target representation information, which not only increases the difficulty of converting high-level semantic features to target mask embedding by a single-layer target model and fails to guide the few-shot learner branch to predict discriminative target representation, but also weakens the target spatio-temporal information that can be retrieved by the query in the transduction branch, thus limiting the representational capability of the generated target-aware feature. 

To tackle this issue, we propose the discriminative label generation module to provide discriminative target feature for both branches. Specifically, our DLGM differs from the JOINT architecture in two points. First, instead of employing the mask of the target object as the input of the label encoder $G_{\theta}$, we propose to adopt the background-filtered frame $\hat{I}$, which can be obtained by simply element-wise multiplying the frame and the corresponding mask. Compared with the single-channel mask, the three-channel BFF can provide not only the target position cues like the mask, but also more representation information of the target. Second, since the input of the label encoder $G_{\theta}$ has been changed to the BFF, the simple five-layer CNN, i.e., label encoder, used by the JOINT is incapable of encoding the BFF into target features with strong representation ability. We therefore leverage a powerful yet lightweight transformer \cite{vitae} as the label generator $G_{\theta}$ to encode the BFF and utilize two convolutional heads to generate the two target features $E_{1}$ and $E_{2}$ for the transduction and induction branches, respectively. Besides, following the setting of JOINT, we apply the cosine similarity loss on $E_{1}$ and $E_{2}$ to make them complementary.

\subsection{Adaptive Fusion Module}
\begin{figure}
    \centering
    \includegraphics[width=0.96\linewidth]{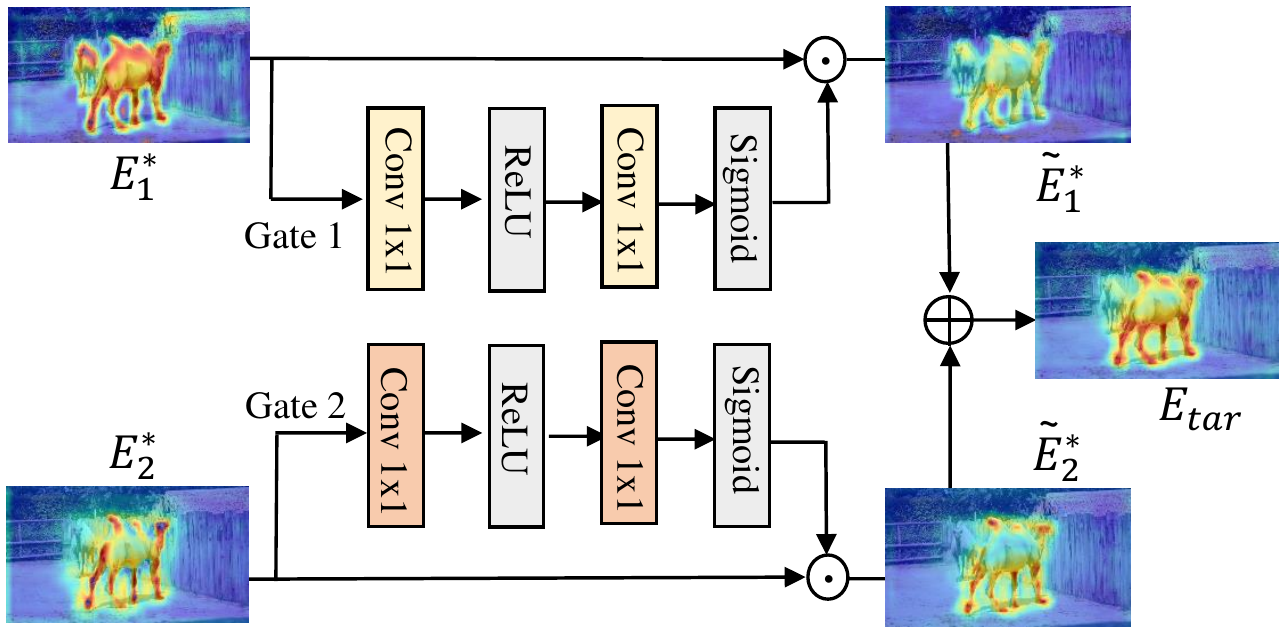}
    \caption{The structure of the adaptive fusion module. Each gate learns an element-wise weight map that reweighs the pixel-level feature representation of each branch before they are fused.}
    \label{afm}
\end{figure}

In the original JOINT framework, the mask embeddings of the two branches are fused by directly pixel-wise summation. Although the cosine similarity loss on $E_{1}$ and $E_{2}$ can indirectly guide the two branches to generate complementary target features, it cannot prevent one branch from becoming dominant and overwhelming the representation of the other branch, and may lead to less discriminative and robust target representation. In order to address this issue, we propose the adaptive fusion module. Our AFM provides a gate for each branch, which learns an element-wise weight map based on the output target feature to reweigh each pixel representation of the feature, thereby adaptively controlling the amount of target information that flows to the fused feature. The AFM is illustrated in Fig. \ref{afm} and can be mathematically described as follows:
\begin{equation}\label{AFM}
\begin{aligned}
	&\widetilde{E}_{1}^{*} = \gamma_{1}\left(E_{i}^{*}\right) \odot {E}_{1}^{*} , \quad \widetilde{E}_{2}^{*} = \gamma_{2}\left(E_{2}^{*}\right) \odot {E}_{2}^{*},\\
	& E_{tar} = \widetilde{E}_{1}^{*} + \widetilde{E}_{2}^{*},
\end{aligned}
\end{equation}
where $\odot$ indicates element-wise multiplication and $\gamma_{1}$ and $\gamma_{2}$ are the two learnable gates, each of them is a two-layer perceptron, which is implemented with a 1$\times$1 convolution layer followed by ReLU and a 1$\times1$ convolution followed by sigmoid function. As shown in Fig. \ref{afm}, through the adaptive fusion module, the original target features of the two branches are reweighed while the distractor representation can also be suppressed. It promotes learning a more robust and discriminative fused target feature $E_{tar}$ for the subsequent decoder.

\subsection{Offline Training}
To simulate the inference procedure of our model, we train the entire framework end-to-end on mini-sequences $V = \{(\mathbf{I}_t, y_{t})\}_{t=0}^{Q}$ of length $Q$=4, in which the samples are randomly sampled from the same video sequence in temporal order. During training, the memory bank is initialized with the first frame-mask pair $\mathbf{M}_0 = \{(\mathbf{x}_0, \hat{I}_{0})\}$, which is used to learn the initial target model parameters $\tau_{0} = A_{\theta}(\mathbf{M}_{0},0)$ by performing 5 steepest descent iterations and implement the matching process in the transformer branch. In subsequent frame, $\tau_{t} = A_{\theta}(\mathbf{M}_{t},\tau_{t-1})$  is updated after 2 iterations. The final mask prediction $\widetilde{y}_{t}$ and the corresponding frame are added to the memory bank $\mathcal{M}_{t}=\mathcal{M}_{t-1} \cup\left\{\left(x_{t}, \hat{I}_{t}\right)\right\}$, where $\hat{I}_{t}$ is the generated BFF. The training loss is the weighted sum of the segmentation loss and the cosine similarity loss, which is the same as JOINT \cite{mao2021joint}.

\subsection{Online Inference}
During inference, a memory bank $\textbf{M}$ is maintained to store the sampled past frames and corresponding masks. Given a test video sequence $V = \{\mathbf{I}_t\}_{t=0}^{Q}$ along with the first annotation $y_{0}$, we first initialize the memory bank with $\mathbf{M}_0 = \{(\mathbf{x}_0, \hat{I}_{0})\}$, where $\mathbf{x}_0 = F_{\theta}(\mathbf{I}_0)$ is the deep feature extracted from $\mathbf{I}_0$ by the backbone network $F_{\theta}$, and $\hat{\mathbf{I}}_{0}$ is the BFF. Then the initial sample is used to learn the target model in the few-shot learner and realize the target information retrieval process in the transformer to perform object segmentation on subsequent frames. To better leverage the temporal information, we sample the past frames at the interval of $T$ = 5 and added to the memory along with the predicted segmentation mask. Except for the first frame-mask pair, the oldest sample will be remove to ensure that the size of the memory does not exceed 20. Additionally, for the online few-shot learning process, our few-shot learner employs 20 iterations in the first frame with a zero initialization $\tau^{0}$ = 0 and 3 iterations in each subsequent sample in the memory to update $\tau_{t} = A_{\theta}(M_{t},\tau_{t-1})$. Before feeding into the model, each frame in the sequence is processed by first cropping a patch that is 5 times larger than the previous estimate of target, while ensuring the maximal size to be equal to the image itself, and then the cropped patch is resized to 832 $\times$ 480. If a video sequence contains multiple objects, each of them is processed independently and the predicted masks are merged together using soft-aggregation \cite{RGMP}.

\section{Experiments}
\subsection{Datasets and Evaluation Metrics}
Our proposed model is evaluated on three benchmark datasets, namely DAVIS 2017 val set \cite{DAVIS2017}, YouTube-VOS val sets of 2018 and 2019 versions \cite{YouTube}. The DAVIS 2017 val set has 30 test video sequences with instance-level annotations. The YouTube-VOS 2018 val set includes 474 videos and the 2019 version is extended to 507 videos with more videos and annotations. Among these videos, some videos contain only a single object, while others have multiple target objects for segmentation. Compared with DAVIS 2017 val set, the YouTube-VOS val sets are more challenging in terms of scenes and video numbers. Besides, for the YouTube-VOS val sets, only the first frame annotation of the target objects is provided in each video, and we must upload the generated segmentation results to the official server for evaluation.

For the DAVIS datasets, we follow its standard protocol, where the Jaccard index ($\mathcal{J}$) measures the region similarity, the $\mathcal{F}$ score indicates the contour accuracy, and $\mathcal{J} \& \mathcal{F}$ is the mean of them. For YouTube-VOS comparison, we calculate $\mathcal{J}$ and $\mathcal{F}$ scores for classes included in the training set (seen) and the ones that are not (unseen), and the overall score $\mathcal{G}$ is computed as the average over all four scores.

\begin{table*}
\begin{center}
\begin{tabular}{lccccccccc}
\toprule
\multirow{2}{*}{Methods} & \multicolumn{5}{c} {YouTube-VOS 2018} & \multicolumn{3}{c} {DAVIS 2017}\\
\cmidrule(lr){2-6} \cmidrule(lr){7-9}
& $\mathcal{G}$ & $\mathcal{J}_{\text{seen}}$ & $\mathcal{F}_{\text{seen}}$ & $\mathcal{J}_{\text{unseen}}$ & $\mathcal{F}_{\text{unseen}}$ &$\mathcal{J \& F}$ & $\mathcal{J}_{\mathcal{M}}$ & $\mathcal{F}_{M}$ & Time (t/s) \\
\midrule
OSVOS \cite{OSVOS} & 58.8 & 59.8 & 60.5 & 54.2 & 60.7 & 60.3 & 56.7  & 63.9 & 9 \\
TVOS \cite{TVOS} & 67.8 & 67.1 & 69.4 & 63.0 & 71.6 & 72.3 & 69.9 & 74.7 & 0.03 \\
GC\cite{GC} & 73.2 & 72.6 & 75.6 & 68.9 & 75.7 & 71.4 & 69.3 & 73.5 & 0.04 \\
Swift\cite{Swift} & 77.8 & 77.8 & 81.8 & 72.3 & 79.5 & 81.8 & 79.2 &  84.3 & 0.04  \\
STM \cite{STM} & 79.4 & 79.7 & 84.2 & 72.8 & 80.9 & 81.8 & 79.2 &  84.3 & 0.12 \\
AFB-URR \cite{liang2020video} & 79.6 & 78.8 & 83.1 & 74.1 & 82.6 & 74.6 & 73.0 & 76.1 & 0.25 \\
GIEL \cite{GIEL} & 80.6 & 80.7 & 85.0 & 75.0 & 81.9 & 82.7 & 80.2 &  85.3 & 0.16  \\
EGMN \cite{EGMN} & 80.2 & 80.7 & 85.1 & 74.0 & 80.9 & 82.8 & 80.2 & 85.2 & 0.20 \\
SITVOS \cite{SITVOS} & 81.3 & 79.9 & 84.3 & 76.4 & 84.4 & 83.5 & 80.4 & 86.5 & 0.10 \\
KMN \cite{KMN} & 81.4 & 81.4 & 85.6 & 75.3 & 83.3 & 84.2 & 80.8 & 87.5 & 0.12 \\
RMNet \cite{RMNet} & 81.5 & 82.1 & 85.7 & 75.7 & 82.4 & 83.5 & 81.0  & 86.0 & 0.10 \\
LWL \cite{LWL} & 81.5 & 80.4 & 84.9 & 76.4 & 84.4 & 81.6 & 79.1 & 84.1 & 0.08  \\
CFBI+ \cite{CFBI} & 82.0 & 81.2 & 86.0 & 76.2 & 84.6 & 82.9 & 80.1 &  85.7 & 0.16 \\
LCM \cite{LCM} & 82.0 & 82.2 & 86.7 & 75.7 & 83.4 & 83.5 & 80.5  & 86.5 & 0.12 \\
HMMN \cite{HMMN} & 82.6 & 82.1 & 87.0 & 76.8 & 84.6 & 84.7 & 81.9 & 87.5 & 0.10 \\
SWEM \cite{SWEM} & 82.8 & 82.4 & 86.9 & 77.1 & 85.0 & 84.3 & 81.2 & 87.4 & 0.04 \\
STCN \cite{STCN} & 83.0 & 81.9 & 86.5 & 77.9 & 85.7 & \textbf{85.4} & 82.2 & 88.6 & 0.06 \\
AOT-L \cite{AOT} & 83.7 & 82.5 & 87.5 & 77.9 & 86.7 & 83.0 & 80.3 & 85.7 & 0.13 \\
\midrule
JOINT \cite{mao2021joint} & 83.1 & 81.5 & 85.9 & 78.7 & 86.5 & 83.5 & 80.8  & 86.2 & 0.10 \\
\textbf{LLB} & \textbf{83.8} & 82.1 & 87.0 & 79.1 & 87.0 & \textbf{84.6} & 81.5 & 87.7 & 0.12 \\
\bottomrule
\end{tabular}
\end{center}
\caption{Quantitative comparison on the YouTube-VOS 2018 and DAVIS 2017 validation sets. t/s: second per frame.}
\label{tab1}
\end{table*}

\subsection{Implementation Details}
Following the setting of JOINT, we choose the ResNet-50 \cite{Resnet}, which is initialized with Mask R-CNN \cite{maskrcnn} weights, as the backbone. The ViTAE-6M \cite{vitae} initialized with ImageNet pre-trained weights is used as the lightweight encoder of the DLGM. The features of stride 16 in both backbone and DLGM are selected as input of the two branches, in which the feature channels are first reduced from 1024 to 512 using an additional convolutional layer, thus the input channel and output channel in both branches are $C=512$ and $D=32$, respectively. For a fair comparison, we use the same decoder $D_{\theta}$ as JOINT.

Considering the transformer branch and the DLGM in our framework, we follow the widely used training protocol in the matching based methods \cite{STM, AOT}. We first pre-train the model on the synthetic video sequence generated from static image dataset COCO \cite{COCO} and then finetune the model on the DAVIS 2017 and YouTube-VOS 2019 training sets. The whole training process contains 140K iterations with a batch size of 32, where the first 50K iterations is the pre-training stage and the rest 90K iterations is the finetuning stage. The model is trained on 4 Nvidia A100 GPUs and tested on a V100 GPU in Pytorch. Like most of the SVOS approaches, we evaluate the running speed of our model on the single-object video sequence. Code is available at \url{https://github.com/ViTAE-Transformer/VOS-LLB}.

\begin{table}[htbp]
	\footnotesize
	\begin{center}
		\begin{tabular*}{\hsize}{@{}@{\extracolsep{\fill}}lccccc@{}}
			\toprule[1.0pt]
		    Methods & $\mathcal{G}$ & $\mathcal{J}_{\text{seen}}$ & $\mathcal{F}_{\text{seen}}$ & $\mathcal{J}_{\text{unseen}}$ & $\mathcal{F}_{\text{unseen}}$ \\
			\midrule
			STM  & 79.2 & 79.6 & 83.6 & 73.0 & 80.6 \\
			KMN  & 80.0 & 80.4 & 84.5 & 73.8 & 81.4 \\
			LWL  & 81.0 & 79.6 & 83.8 & 76.4 & 84.2 \\
			CFBI  & 81.0 & 80.6 & 85.1 & 75.2 & 83.0 \\
			HMMN  & 82.5 & 81.7 & 86.1 & 77.3 & 85.0 \\
			STCN & 82.7 & 81.1 & 85.4 & 78.2 & 85.9  \\
			RDE-VOS  & 82.8 & 82.4 & 86.9 & 77.1 & 85.0  \\
			AOT-L & 83.6 & 82.2 & 86.9 & 78.3 & 86.9 \\
			\midrule
			JOINT & 82.8 & 80.8 & 84.8 & 79.0 & 86.6 \\
			\textbf{LLB} & \textbf{83.6} & 81.7 & 86.5 & 79.2 & 87.0  \\
			\midrule[1.0pt]
		\end{tabular*}
	\end{center}
	\caption{Comparison on the YouTube-VOS 2019 val set.}
	\label{tab2}
\end{table}

\subsection{Comparison with State-of-the-art Methods}

\noindent\textbf{YouTube-VOS.} Here, we compare the quantitative results of our proposed LLB model on the challenging YouTube-VOS val sets with the state-of-the-art methods. For the YouTube-VOS 2018 val set, as presented in Table~\ref{tab1}, our model achieves the average scores of 83.8\% $\mathcal{G}$, which outperforms the state-of-the-art methods, such as SWEM \cite{SWEM}, STCN \cite{STCN} and AOT \cite{AOT}. Particularly, compared with the baseline JOINT, our LLB surpasses it in all four metrics of seen and unseen object categories, and is 0.7\% higher than it on the overall accuracy. We report the quantitative comparison on the YouTube-VOS 2019 val set in Table~\ref{tab2}, and our proposed LLB also achieves state-of-the-art overall accuracy of 83.6\% $\mathcal{G}$, which exceeds the baseline JOINT by 0.8\% and performs better than the latest RDE-VOS \cite{RDEVOS} and HMMN \cite{HMMN}. Moreover, the joint learning based methods, e.g., JOINT and our LLB, have similar performance with the matching based ones, such as HMMN and STCN, on the seen categories, while the joint learning based methods perform much better on the unseen categories. We believe the reason for this is that the online few-shot learner branch adapts the model better to new object categories during inference and improves the generalization ability of the whole model.

\begin{figure*}[t]
    \centering
    \includegraphics[width=0.96\linewidth]{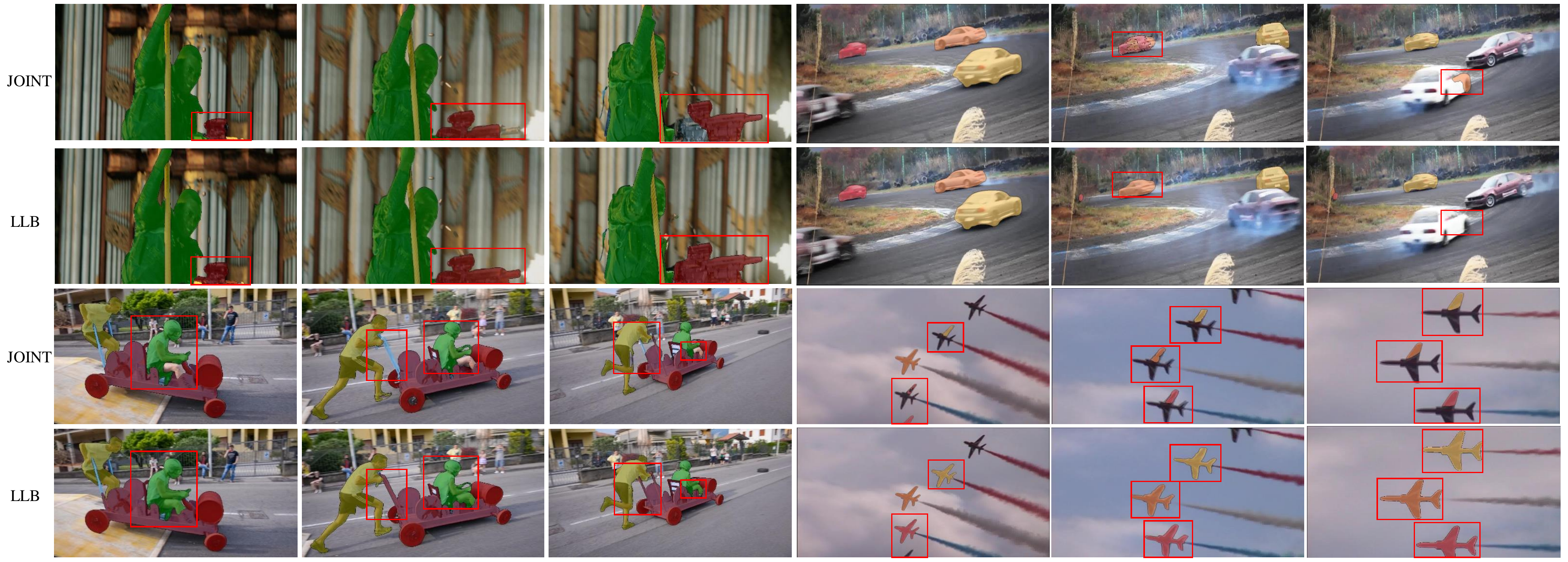}
    \caption{Visual results comparison between JOINT and LLB. We highlight the differences with red boxes.}
    \label{vis_res}
\end{figure*}

\noindent\textbf{DAVIS.} We also evaluate the performance of our approach on the DAVIS 2017 val set. Since the DAVIS 2017 val set has fewer test videos and relatively easy scenes, the latest approaches achieve very similar performance. The quantitative comparison with other state-of-the-art methods are reported in Table~\ref{tab1}. Our approach achieves the competitive performance of 84.6\% $\mathcal{J} \& \mathcal{F}$ which significantly improves the baseline JOINT by 1.1\% and also outperforms the other latest comparison methods, e.g., RED-VOS \cite{RDEVOS} and LCM \cite{LCM}. Moreover, the running speed of our model is also competitive compared with other state-of-the-art approaches.

Some qualitative comparisons between our LLB and the baseline JOINT are presented in Fig. \ref{vis_res}. As we can see that our approach performs better in complicated scenes, e.g., clear details in the boundary area between targets and the mitigation of false segmentation. The significant accuracy improvement over the baseline JOINT and the state-of-the-art performance on the benchmarks demonstrate the effectiveness of our proposed modules and the superiority of our method.

\begin{table} 
	\begin{center}
			\begin{tabular*}{\hsize}{@{}@{\extracolsep{\fill}}lccccc@{}}
				\toprule[1.0pt]
				Version & DLGM & AFM & $\mathcal{J} \& \mathcal{F}$ & $\mathcal{J}$ & $\mathcal{F}$ \\
				\midrule
				Baseline &  &  & 83.5 & 80.8 & 86.2\\
				LLB-1  & \checkmark &  & 84.2 & 81.2 & 87.1\\
				LLB-2  &  & \checkmark & 83.8 & 81.0 & 86.6\\
				LLB    & \checkmark & \checkmark & \textbf{84.6} & 81.5 & 87.7\\
				\bottomrule[1.0pt]
			\end{tabular*}
	\end{center}
	\caption{Ablation study of each component on DAVIS 2017 val dataset. Baseline denotes the JOINT method.}
	\label{ablation}
\end{table}

\begin{table} 
	\begin{center}
			\begin{tabular*}{\hsize}{@{}@{\extracolsep{\fill}}lccccc@{}}
				\toprule[1.0pt]
				Version & Input & Encoder & $\mathcal{J} \& \mathcal{F}$ & $\mathcal{J}$ & $\mathcal{F}$ \\
				\midrule
				Baseline  & mask & 5-layer CNN & 83.8 & 81.0 & 86.6\\
				LLB-3  & BFF & 5-layer CNN & 84.1 & 81.2 & 87.1\\
				LLB-4  & mask & ViTAE-6M & 83.9 & 81.0 & 86.8\\
				LLB    & BFF & ViTAE-6M & \textbf{84.6} & 81.5 & 87.7\\
				\bottomrule[1.0pt]
			\end{tabular*}
	\end{center}
	\caption{Ablation study of the DLGM on DAVIS 2017 val dataset. Input denotes the input of DLGM. Encoder denotes the generator to encode the input. 5-layer Conv denotes the original encoder in JOINT. Baseline means the JOINT with AFM.}
	\label{dlg}
\end{table}

\subsection{Ablation Study}
\noindent\textbf{Module Ablation.} To verify the effectiveness of each module in our model, we perform ablation studies on the DAVIS 2017 val set. As shown in Table~\ref{ablation}, on the base of baseline JOINT, if we only apply the DLGM, the performance of the LLB-1 is significantly improved by 0.7\%, and if only the AFM is deployed on the baseline, the LLB-2 model brings 0.3\% performance gain. Specifically, when we combine the two module together, the complete LLB achieves the best 84.6\% $\mathcal{J} \& \mathcal{F}$. These experiments prove the effectiveness of each module and also show that putting them together performs better.

\noindent\textbf{DLGM.} To investigate the influence of different components in the discriminative label generation module, we conduct several ablation experiments on the DAVIS 2017 val set. As reported in Table~\ref{dlg}, the baseline is the JOINT with AFM, in which the input of the label encoder is the one-channel mask and the encoder is a 5-layer CNN. LLB-3 takes the BFF as input and improves over the baseline by 0.3\%, which suggests that it is beneficial to use the BFF to provide more target information. LLB-4 only replaces the 5-layer CNN with the ViTAE-6M while the improvement is limited, which means the impact of a more powerful encoder is marginal when mask is used as input. The full version LLB takes advantage of the BFF and lightweight yet powerful encoder and delivers the best performance.

\section{Conclusion}
In this work, we introduce a new joint learning based framework for SVOS task, which emphasizes learning to learn better target features. Two key designs have been proposed to improve the target representation. Specifically, the discriminative label generation module uses a lightweight transformer to encode the background-filtered frame, providing discriminative guidance to the two branches to generate better target representation. The adaptive fusion module learns a gate for each branch to adaptively reweigh the feature representation and then fuses the target features in both branches to produce a more robust and discriminative target representation. Experimental results on several public benchmark datasets show that the proposed method outperforms representative SVOS methods and achieves state-of-the-art performance.


\bibliography{aaai23}

\begin{thebibliography}{32}
\providecommand{\natexlab}[1]{#1}

\bibitem[{Bhat et~al.(2020)Bhat, Lawin, Danelljan, Robinson, Felsberg, Gool,
  and Timofte}]{LWL}
Bhat, G.; Lawin, F.~J.; Danelljan, M.; Robinson, A.; Felsberg, M.; Gool, L.~V.;
  and Timofte, R. 2020.
\newblock Learning What to Learn for Video Object Segmentation.
\newblock In \emph{ECCV}, 777--794.

\bibitem[{Caelles et~al.(2017)Caelles, Maninis, Pont{-}Tuset,
  Leal{-}Taix{\'{e}}, Cremers, and Gool}]{OSVOS}
Caelles, S.; Maninis, K.; Pont{-}Tuset, J.; Leal{-}Taix{\'{e}}, L.; Cremers,
  D.; and Gool, L.~V. 2017.
\newblock One-Shot Video Object Segmentation.
\newblock In \emph{CVPR}, 5320--5329.

\bibitem[{Chen et~al.(2020)Chen, Li, Yuan, Yu, Shen, and Qi}]{SAT}
Chen, X.; Li, Z.; Yuan, Y.; Yu, G.; Shen, J.; and Qi, D. 2020.
\newblock State-Aware Tracker for Real-Time Video Object Segmentation.
\newblock In \emph{CVPR}, 9381--9390.

\bibitem[{Cheng, Tai, and Tang(2021)}]{STCN}
Cheng, H.~K.; Tai, Y.-W.; and Tang, C.-K. 2021.
\newblock Rethinking Space-Time Networks with Improved Memory Coverage for
  Efficient Video Object Segmentation.
\newblock In \emph{NeurIPS}, 11781--11794.

\bibitem[{Dosovitskiy et~al.(2020)Dosovitskiy, Beyer, Kolesnikov, Weissenborn,
  Zhai, Unterthiner, Dehghani, Minderer, Heigold, Gelly et~al.}]{vit}
Dosovitskiy, A.; Beyer, L.; Kolesnikov, A.; Weissenborn, D.; Zhai, X.;
  Unterthiner, T.; Dehghani, M.; Minderer, M.; Heigold, G.; Gelly, S.; et~al.
  2020.
\newblock An Image is Worth 16x16 Words: Transformers for Image Recognition at
  Scale.
\newblock In \emph{ICLR}.

\bibitem[{Ge, Lu, and Shen(2021)}]{GIEL}
Ge, W.; Lu, X.; and Shen, J. 2021.
\newblock Video Object Segmentation Using Global and Instance Embedding
  Learning.
\newblock In \emph{CVPR}, 16836--16845.

\bibitem[{He et~al.(2017)He, Gkioxari, Doll{\'a}r, and Girshick}]{maskrcnn}
He, K.; Gkioxari, G.; Doll{\'a}r, P.; and Girshick, R. 2017.
\newblock Mask R-cnn.
\newblock In \emph{ICCV}, 2961--2969.

\bibitem[{He et~al.(2016)He, Zhang, Ren, and Sun}]{Resnet}
He, K.; Zhang, X.; Ren, S.; and Sun, J. 2016.
\newblock Deep Residual Learning for Image Recognition.
\newblock In \emph{CVPR}, 770--778.

\bibitem[{Hu et~al.(2021)Hu, Zhang, Zhang, Pan, Xu, and Jin}]{LCM}
Hu, L.; Zhang, P.; Zhang, B.; Pan, P.; Xu, Y.; and Jin, R. 2021.
\newblock Learning Position and Target Consistency for Memory-based Video
  Object Segmentation.
\newblock In \emph{CVPR}, 4144--4154.

\bibitem[{Lan et~al.(2022)Lan, Zhang, He, and Zhang}]{SITVOS}
Lan, M.; Zhang, J.; He, F.; and Zhang, L. 2022.
\newblock Siamese Network with Interactive Transformer for Video Object
  Segmentation.
\newblock In \emph{AAAI}, 1228--1236.

\bibitem[{Li et~al.(2022)Li, Hu, Xiong, Zhang, Pan, and Liu}]{RDEVOS}
Li, M.; Hu, L.; Xiong, Z.; Zhang, B.; Pan, P.; and Liu, D. 2022.
\newblock Recurrent Dynamic Embedding for Video Object Segmentation.
\newblock In \emph{CVPR}, 1332--1341.

\bibitem[{Li, Shen, and Shan(2020)}]{GC}
Li, Y.; Shen, Z.; and Shan, Y. 2020.
\newblock Fast Video Object Segmentation Using the Global Context Module.
\newblock In \emph{ECCV}, 735--750.

\bibitem[{Liang et~al.(2020)Liang, Li, Jafari, and Chen}]{liang2020video}
Liang, Y.; Li, X.; Jafari, N.; and Chen, J. 2020.
\newblock Video Object Segmentation with Adaptive Feature Bank and
  Uncertain-Region Refinement.
\newblock In \emph{NeurIPS}, 3430--3441.

\bibitem[{Lin et~al.(2014)Lin, Maire, Belongie, Hays, Perona, Ramanan,
  Doll{\'{a}}r, and Zitnick}]{COCO}
Lin, T.; Maire, M.; Belongie, S.~J.; Hays, J.; Perona, P.; Ramanan, D.;
  Doll{\'{a}}r, P.; and Zitnick, C.~L. 2014.
\newblock Microsoft {COCO:} Common Objects in Context.
\newblock In \emph{ECCV}, 740--755.

\bibitem[{Lin et~al.(2022)Lin, Yang, Li, Wang, Yuan, Jiang, and Liu}]{SWEM}
Lin, Z.; Yang, T.; Li, M.; Wang, Z.; Yuan, C.; Jiang, W.; and Liu, W. 2022.
\newblock SWEM: Towards Real-Time Video Object Segmentation with Sequential
  Weighted Expectation-Maximization.
\newblock In \emph{CVPR}, 1362--1372.

\bibitem[{Lu et~al.(2021)Lu, Cole, Dekel, Zisserman, Freeman, and
  Rubinstein}]{lu2021omnimatte}
Lu, E.; Cole, F.; Dekel, T.; Zisserman, A.; Freeman, W.~T.; and Rubinstein, M.
  2021.
\newblock Omnimatte: Associating Objects and Their Effects in Video.
\newblock In \emph{CVPR}, 4507--4515.

\bibitem[{Lu et~al.(2020)Lu, Wang, Martin, Zhou, Shen, and Luc}]{EGMN}
Lu, X.; Wang, W.; Martin, D.; Zhou, T.; Shen, J.; and Luc, V.~G. 2020.
\newblock Video Object Segmentation with Episodic Graph Memory Networks.
\newblock In \emph{ECCV}, 661--679.

\bibitem[{Luiten, Voigtlaender, and Leibe(2018)}]{PReMVOS}
Luiten, J.; Voigtlaender, P.; and Leibe, B. 2018.
\newblock PReMVOS: Proposal-Generation, Refinement and Merging for Video Object
  Segmentation.
\newblock In \emph{ACCV}, 565--580.

\bibitem[{Mao et~al.(2021)Mao, Wang, Zhou, and Li}]{mao2021joint}
Mao, Y.; Wang, N.; Zhou, W.; and Li, H. 2021.
\newblock Joint Inductive and Transductive Learning for Video Object
  Segmentation.
\newblock In \emph{ICCV}, 9670--9679.

\bibitem[{Oh et~al.(2018)Oh, Lee, Sunkavalli, and Kim}]{RGMP}
Oh, S.~W.; Lee, J.; Sunkavalli, K.; and Kim, S.~J. 2018.
\newblock Fast Video Object Segmentation by Reference-Guided Mask Propagation.
\newblock In \emph{CVPR}, 7376--7385.

\bibitem[{Oh et~al.(2019)Oh, Lee, Xu, and Kim}]{STM}
Oh, S.~W.; Lee, J.; Xu, N.; and Kim, S.~J. 2019.
\newblock Video Object Segmentation Using Space-Time Memory Networks.
\newblock In \emph{ICCV}, 9225--9234.

\bibitem[{Pont{-}Tuset et~al.(2017)Pont{-}Tuset, Perazzi, Caelles, Arbelaez,
  Sorkine{-}Hornung, and Gool}]{DAVIS2017}
Pont{-}Tuset, J.; Perazzi, F.; Caelles, S.; Arbelaez, P.; Sorkine{-}Hornung,
  A.; and Gool, L.~V. 2017.
\newblock The 2017 {DAVIS} Challenge on Video Object Segmentation.
\newblock abs/1704.00675.

\bibitem[{Robinson et~al.(2020)Robinson, Lawin, Danelljan, Khan, and
  Felsberg}]{FRTM}
Robinson, A.; Lawin, F.~J.; Danelljan, M.; Khan, F.~S.; and Felsberg, M. 2020.
\newblock Learning Fast and Robust Target Models for Video Object Segmentation.
\newblock In \emph{CVPR}, 7404--7413.

\bibitem[{Seong, Hyun, and Kim(2020)}]{KMN}
Seong, H.; Hyun, J.; and Kim, E. 2020.
\newblock Kernelized Memory Network for Video Object Segmentation.
\newblock In \emph{ECCV}, 629--645.

\bibitem[{Seong et~al.(2021)Seong, Oh, Lee, Lee, Lee, and Kim}]{HMMN}
Seong, H.; Oh, S.~W.; Lee, J.-Y.; Lee, S.; Lee, S.; and Kim, E. 2021.
\newblock Hierarchical Memory Matching Network for Video Object Segmentation.
\newblock In \emph{ICCV}, 12889--12898.

\bibitem[{Wang et~al.(2021)Wang, Jiang, Ren, Hu, and Bai}]{Swift}
Wang, H.; Jiang, X.; Ren, H.; Hu, Y.; and Bai, S. 2021.
\newblock SwiftNet: Real-time Video Object Segmentation.
\newblock In \emph{CVPR}, 1296--1305.

\bibitem[{Xie et~al.(2021)Xie, Yao, Zhou, Zhang, and Sun}]{RMNet}
Xie, H.; Yao, H.; Zhou, S.; Zhang, S.; and Sun, W. 2021.
\newblock Efficient Regional Memory Network for Video Object Segmentation.
\newblock In \emph{CVPR}, 1286--1295.

\bibitem[{Xu et~al.(2018)Xu, Yang, Fan, Yang, Yue, Liang, Price, Cohen, and
  Huang}]{YouTube}
Xu, N.; Yang, L.; Fan, Y.; Yang, J.; Yue, D.; Liang, Y.; Price, B.~L.; Cohen,
  S.; and Huang, T.~S. 2018.
\newblock YouTube-VOS: Sequence-to-Sequence Video Object Segmentation.
\newblock In \emph{ECCV}, 603--619.

\bibitem[{Xu et~al.(2021)Xu, Zhang, Zhang, and Tao}]{vitae}
Xu, Y.; Zhang, Q.; Zhang, J.; and Tao, D. 2021.
\newblock ViTAE: Vision Transformer Advanced by Exploring Intrinsic Inductive
  Bias.
\newblock In \emph{NeurIPS}, 28522--28535.

\bibitem[{Yang, Wei, and Yang(2021{\natexlab{a}})}]{AOT}
Yang, Z.; Wei, Y.; and Yang, Y. 2021{\natexlab{a}}.
\newblock Associating Objects with Transformers for Video Object Segmentation.
\newblock In \emph{NeurIPS}, 2491--2502.

\bibitem[{Yang, Wei, and Yang(2021{\natexlab{b}})}]{CFBI}
Yang, Z.; Wei, Y.; and Yang, Y. 2021{\natexlab{b}}.
\newblock Collaborative Video Object Segmentation by Multi-Scale
  Foreground-Background Integration.
\newblock \emph{TPAMI}, 4701--4712.

\bibitem[{Zhang et~al.(2020)Zhang, Wu, Peng, and Lin}]{TVOS}
Zhang, Y.; Wu, Z.; Peng, H.; and Lin, S. 2020.
\newblock A Transductive Approach for Video Object Segmentation.
\newblock In \emph{CVPR}, 6949--6958.

\end{thebibliography}

\end{document}